\documentclass[10pt, conference]{IEEEtran}

\usepackage{balance, cite}

\usepackage{graphicx, subfigure}

\usepackage[cmex10]{amsmath}
\usepackage{amsthm}

\usepackage{xcolor, url}

\newcommand{\up}[1]{\raisebox{1.5ex}[0pt]{#1}}

\begin{document}
%
\thispagestyle{empty}
\onecolumn
\linespread{1.2}\selectfont{}
{\noindent\Huge IEEE Copyright Notice}\\[1pt]

{\noindent\large Copyright (c) 2016 IEEE

\noindent Personal use of this material is permitted. Permission from IEEE must be obtained for all other uses, in any current or future media, including reprinting/republishing this material for advertising or promotional purposes, creating new collective works, for resale or redistribution to servers or lists, or reuse of any copyrighted component of this work in other works.}\\[1em]

{\noindent\Large Accepted to be published in: 2016 29th SIBGRAPI Conference on Graphics, Patterns and Images (SIBGRAPI'16), October 4--7, 2016.}\\[1in]

{\noindent\large Cite as:}\\[1pt]

{\setlength{\fboxrule}{1pt}
 \fbox{\parbox{0.7\textwidth}{L. A. Duarte, O. A. B. Penatti and J. Almeida, ``Bag of Genres for Video Retrieval,'' in \emph{2016 29th SIBGRAPI Conference on Graphics, Patterns and Images (SIBGRAPI)}, S\~{a}o Jos\'{e} dos Campos, Brazil, 2016, pp. 257-264, doi: 10.1109/SIBGRAPI.2016.043}}}\\[1in]
 
{\noindent\large BibTeX:}\\[1pt]

{\setlength{\fboxrule}{1pt}
 \fbox{\parbox{0.85\textwidth}{
 @InProceedings\{SIBGRAPI\_2016\_Duarte,
 
 \begin{tabular}{lll}
  & author    & = \{L. A. \{Duarte\} and O. A. B. \{Penatti\} and J. \{Almeida\}\},\\
			   
  & title     & = \{Bag of Genres for Video Retrieval\},\\
			   
  & pages     & = \{257--264\},\\
  
  & booktitle & = \{2016 29th {SIBGRAPI} Conference on Graphics, Patterns and Images ({SIBGRAPI})\},\\
  
  & address   & = \{S\~{a}o Jos\'{e} dos Campos, Brazil\},\\
  
  & month     & = \{October 4--7\},\\
  
  & year      & = \{2016\},\\
  
  & publisher & = \{\{IEEE\}\},\\
  
  & doi       & = \{10.1109/SIBGRAPI.2016.043\},\\
  \end{tabular}
  
\}
 }}}

\twocolumn
\linespread{1}\selectfont{}
\clearpage

%
\title{Bag of Genres for Video Retrieval}

\newif\iffinal
\finaltrue
\newcommand{\jemsid}{135}


\iffinal
  \author{%
	\IEEEauthorblockN{Leonardo A. Duarte$^1$, Ot\'{a}vio A. B. Penatti$^2$, and Jurandy Almeida$^1$}\\
	\IEEEauthorblockA{%
	  $^1$Institute of Science and Technology\\
	  Federal University of S\~{a}o Paulo -- UNIFESP\\
	  12247-014, S\~{a}o Jos\'{e} dos Campos, SP -- Brazil\\
	  Email: \small\texttt{\{leonardo.assuane, jurandy.almeida\}@unifesp.br}\\[1ex]
	  $^2$Advanced Technologies\\
	  SAMSUNG Research Institute\\
	  13097-160, Campinas, SP -- Brazil\\
	  Email: \small\texttt{o.penatti@samsung.com}}
  }
\else
  \author{Sibgrapi paper ID: \jemsid \\ }
\fi

%


\maketitle

\begin{abstract}
Often, videos are composed of multiple concepts or even genres. 
For instance, news videos may contain sports, action, nature, etc. 
Therefore, encoding the distribution of such concepts/genres in a 
compact and effective representation is a challenging task. In this 
sense, we propose the Bag of Genres representation, which is based 
on a visual dictionary defined by a genre classifier. Each visual 
word corresponds to a region in the classification space. The Bag 
of Genres video vector contains a summary of the activations of each 
genre in the video content. We evaluate the proposed method for video 
genre retrieval using the dataset of MediaEval Tagging Task of 2012 
and for video event retrieval using the EVVE dataset. Results show 
that the proposed method achieves results comparable or superior to 
state-of-the-art methods, with the advantage of providing a much more 
compact representation than existing features.

%
\end{abstract}


\IEEEpeerreviewmaketitle



\section{Introduction}
\label{sec:intro}
The retrieval of videos by content is a challenging application, as videos 
may be composed of visually different excepts. For instance, a news video 
can comprise multiple categories, like sports, documentary, health, and 
others. A video retrieval system aiming at retrieving videos with similar 
content should be aware of such property in order to obtain better results.

In this paper, we focus on video retrieval based only on visual 
information. No tags or textual descriptions are considered. One important 
step in this scenario is feature extraction from videos. There are mainly 
two kinds of feature descriptors for videos: descriptors that consider motion 
and descriptors based on isolated frames. Motion-based descriptors usually 
obtain space-time interest points and extract histograms of those local 
points or obtain histogram of motion patterns~\cite{ICIP_2011_Almeida}.
Descriptors based on isolated frames are usually derived from image feature 
extraction. Frames are represented individually and then a pooling function 
can be used to obtain the video feature vector. The advantage of the first 
kind of descriptors is obviously the encoding of transitions between frames.
The advantage of the second kind is the possibility to use the large number 
of descriptors already proposed for image representation.

Regardless of motion, many of the state-of-the-art solutions for feature 
extraction are based on visual dictionaries. Such dictionaries are commonly 
based on local patches, which are semantically poor. Therefore, both kinds 
of descriptors usually present the same property: the video feature vector 
has few semantics from the human perspective. 

In this paper, we present a novel approach for video representation, 
called Bag of Genres (BoG). The proposed 
method is based on dictionaries of genres created from genre classifiers. 
Each visual word in the BoG model is a genre-labeled region of the 
classification space defined by the classifier's model.
The main advantages of the BoG model are the following:
(i) each visual word explicitly contains semantics, which was learned from the labeled data by the genre classifier; 
(ii) the video representation corresponds to an activation vector of its contents to each of the genres in the dictionary, thus having one dimension for each genre; 
and (iii) compact representation, directly related to the number of genres in the dictionary.

We validated the BoG model for video genre retrieval and for video event retrieval.
In the first case, we used the dataset of MediaEval Tagging Task of 2012.
We evaluate the importance of the genre classifier in the model as well as the quality of the BoG representation. 
Although the genre classifier has low accuracy, the BoG model worked well in the experiments. 
The results are comparable to the existing baselines, but BoG is much more compact. 
In the second case, we used our best BoG representation to retrieve videos by 
event on the EVVE dataset. The results in this dataset indicate that the BoG approach 
outperforms state-of-the-art methods.

The remainder of the paper is organized as follows. 
Section~\ref{sec:related} presents related work.
Section~\ref{sec:bogmodel} explains the proposed BoG model.
Section~\ref{sec:results} shows experiments and results, and 
Section~\ref{sec:conclusions} concludes the work indicating possible future work.

\begin{figure*}[!htb]
	\centering
	\includegraphics[width=\textwidth]{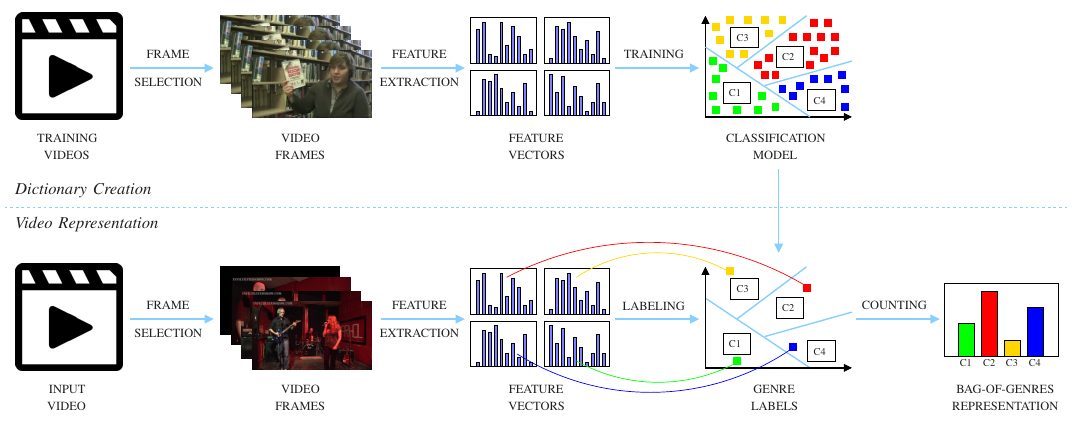}
	\caption{An overview of the Bag-of-Genres model.}
	\label{fig:bogmodel}
\end{figure*}

\section{Related Work}
\label{sec:related}
In this section, we describe related work focusing on works that are 
based on visual dictionaries and works that aim at including semantics 
in the representation.

Many solutions exist in the literature aiming at including semantics in 
the representation. There are techniques in which an image is represented 
as a scale-invariant response map of a large number of pre-trained generic 
object detectors~\cite{NIPS_2010_Li}, which could be seen as a dictionary 
of objects. Poselets have also been used similarly to a dictionary of poses 
for recognizing people poses~\cite{BourdevPoseletsICCV2009}. Labeled 
local patches have also been used for having a dictionary with more 
semantics~\cite{VogelSemanticIJCV2007}. Boureau~et~al.~\cite{CVPR_2010_Boureau} 
also present a way to supervise the dictionary creation. Other approaches 
can also be considered as related to the intention of having dictionaries 
with more meaningful visual words~\cite{AlbaradeiMidLevel2014, 
BhattacharyaConceptDetector2014, PattersonSUN2014, CVPR_2012_Sadanand}

The approach proposed here is closely related to the Bag-of-Scenes 
(BoS) model~\cite{ICMR_2012_Penatti}, in which the video feature 
vector is an activation vector of scenes. As scenes are more semantically 
meaningful than local patches, the BoS feature space is semantically richer. 
Each dimension in the BoS space corresponds to a semantic concept.

The main novelty of BoG in relation to previous works, specially BoS, is 
that we use a genre classifier as visual dictionary. In the BoS model, the 
visual dictionary is based directly on the feature vectors of the scenes.
The advantages of using a classifier is that it better delineates the 
frontiers among visual words and tends to be more robust to feature 
dimensionality. Another advantage is the compact BoG vector, as its 
dimensionality directly corresponds to the number of genres in the problem.

\section{Bag of Genres}
\label{sec:bogmodel}
In this section, we describe the Bag-of-Genres~(BoG) model for video 
representation. This model is based on a dictionary of genres, in which 
each visual word corresponds to a decision region of the classification 
model defined by a genre classifier. Thus, each video is represented by 
a vector of activations of its frames to each of the genres in the dictionary.

An interesting property of the BoG model is that it relies on elements 
that have more semantics according to the human perception. Traditional 
dictionaries based on local features, like SIFT or STIP, are composed 
of visual words which carry no semantic information, like corners and 
edges~\cite{NIPS_2010_Li}. In the BoG model, as the visual words are 
genre-labeled regions of the classification space, the activation vector 
has one dimension for each genre, making it simple to analyze the presence 
or absence of each genre into a video. 
Another important aspect of using a genre classifier to encode visual 
features, is that the classifier better delineates the feature space 
and classifiers (e.g., Support Vector Machines) deal well with high 
dimensional spaces.

Figure~\ref{fig:bogmodel} shows a flowchart of the stages involved 
in representing video content using the BoG model. On top, we show 
how the visual dictionary is created. At the bottom, we show how this 
codebook is used to represent video content.

The creation of the visual dictionary is performed as follows. Given a 
set of \emph{training videos} with known genre labels, we first discard 
a lot of redundant information, taking only a subset of \emph{video frames}. 
Techniques like sampling at fixed-time intervals or summarization 
methods~\iffinal\cite{ISM_2010_Almeida, PRL_2012_Almeida}\else\cite{PRL_2011_Avila}\fi are examples of 
possibilities for \emph{frame selection}. In this paper, frames were selected 
using the well-known FFmpeg tool\footnote{\url{http://www.ffmpeg.org/} 
(As of May 2016).} in a sampling rate of one frame per second. 
After that, we perform the \emph{feature extraction} from each 
of the selected frames in order to encode their visual content 
into \emph{feature vectors}. Such features can be any, like for 
instance, color histograms, GIST, bags of quantized SIFT features, or even 
features extracted from deep convolutional neural networks~\cite{ICLR_2014_SERMANET}.
Then, those feature vectors and their associated genre labels 
are used as input for \emph{training} a genre classifier. The 
obtained \emph{classification model} represents the dictionary 
of genres used for representing videos.

After creating the visual dictionary, we should represent videos 
according to the dictionary space. Given an \emph{input video}, we 
initially apply \emph{frame selection} and \emph{feature extraction} 
from each frame. After that, the feature vectors of each frame must 
be coded according to the dictionary of genres.
Each feature vector is classified by the genre classifier, 
which predicts a \emph{genre label} for the frame. 
The \emph{labeling} process is analogous to the \emph{coding} step 
of traditional visual dictionaries~\cite{TPAMI_2010_Gemert}. Finally, 
a normalized frequency histogram is obtained by \emph{counting} the 
occurrences of each of the genre labels, forming the \emph{bag-of-genres 
representation} for the input video. Such step can be understood 
as \emph{pooling} the frame genres~\cite{CVPR_2010_Boureau}.

The dimensionality of the bag-of-genres feature space is directly 
related to the number of genres used for training the genre classifier 
during the dictionary creation. Therefore, as in many applications the 
number of genres is small, the bag of genres is usually more compact 
than existing solutions.

\section{Experiments and Results}
\label{sec:results}
We evaluate the BoG model on two challenging tasks: for video genre 
retrieval, using the dataset of MediaEval Tagging Task of 2012; and 
for video event retrieval, using the EVVE dataset. In the following
subsections, we report and discuss the obtained results.

\subsection{Video Genre Retrieval}
\label{sec:results:genre}
Experiments were conducted on a benchmarking dataset 
provided by the MediaEval 2012 organizers for the Genre 
Tagging Task~\cite{MEDIAEVAL_2012_Schmiedeke}. The dataset 
is composed of 14,838 videos (3,288 hours) collected from 
the blip.tv\footnote{\url{http://blip.tv} (As of May 2016).} 
and is divided into a training set of 5,288 videos (36\%) and 
a test set of 9,550 videos (64\%). Those videos are distributed 
among 26 video genre categories assigned by the blip.tv media 
platform, namely (the numbers in brackets are the total number 
of videos): art (530), autos and vehicles (21), business (281), 
citizen journalism (401), comedy (515), conferences and other 
events (247), documentary (353), educational (957), food and 
drink (261), gaming (401), health (268), literature (222), movies 
and television (868), music and entertainment (1148), personal or 
auto-biographical (165), politics (1107), religion (868), school 
and education (171), sports (672), technology (1343), environment 
(188), mainstream media (324), travel (175), video blogging (887), 
web development (116), and default category (2349, which comprises 
videos that cannot be assigned to any of the previous categories). 
The main challenge of this collection is the high diversity of genres, 
as well as the high variety of visual contents within each genre 
category~\cite{MTA_2012_Ionescu, CBMI_2013_Mironica}.

After frame selection (1 per second), the training set has 
3,943,375 frames and the test set has 7,273,996 frames. Different image 
descriptors were evaluated for extracting features from such frames. 
The descriptors for encoding color properties are: Auto Color 
Correlogram (ACC)~\cite{CVPR_1997_Huang}, Color Coherent Vector 
(CCV)~\cite{MM_1996_Pass}, Border/Interior pixel Classification 
(BIC)~\cite{CIKM_2002_Stehling}, and Global Color Histogram 
(GCH)~\cite{IJCV_1991_Swain}. The texture descriptors evaluated 
are: Generic Fourier Descriptor (GFD)~\cite{SPIC_2002_Zhang} and 
Haar-Wavelet Descriptor (HWD)~\cite{SIGGRAPH_1995_Jacobs}. 
For more details regarding those image descriptors, 
please refer to~\cite{JVCIR_2012_Penatti}.

The experiments are divided into three phases. The first one evaluates 
the genre classifier. The second one evaluates the BoG representation 
for video genre retrieval and the last one evaluates the BoG representation 
for video event retrieval. 

\subsubsection{Evaluation of the genre classifier}
\label{sec:results_1}
The evaluation of the genre classifier is important because 
the quality of the final BoG vector depends on the quality 
of this classifier. If the genre classifier classifies 
the frames in wrong genres, the BoG vector will not reflect 
the correct distribution of video genres. It would be similar 
to have a bad coding step in traditional visual dictionaries 
of quantized local features: wrong visual word labels would 
be assigned to image patches, resulting in a bad bag of visual 
words. Therefore, the BoG model depends on a good genre classifier.

To create the visual dictionary, we trained a linear SVM ($c = 1.0$) 
using features extracted from the training videos. The genre (label) of 
each training frame is the same of the video from where it was extracted. 
The training of the genre classifier was based on randomly selecting the 
same number $N$ of frames per genre. We varied $N$ in 100, 500, and 800 
frames per genre. The remaining frames were used for testing. It is worth 
mentioning the amount of frames used in this evaluation: almost 4 million 
of the training videos and more than 7.2 million of the test videos (no 
frames of the test videos were used for training the genre classifier). 
For running SVM, we used the LIBSVM 
package\footnote{\url{http://www.csie.ntu.edu.tw/~cjlin/libsvm/} 
(As of May 2016)}~\cite{TIST_2011_Chang}. 

\begin{figure}[!htb]
	\centering
	\includegraphics[width=0.45\textwidth]{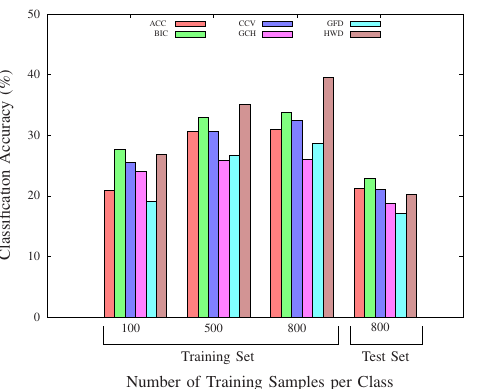}
	\caption{Evaluation of the genre classifier. All descriptors 
	         generated low discriminating genre classifiers (accuracy 
	         below 50\%), creating a challenging scenario for the BoG model.}
	\label{fig:accuracy}
\end{figure}

Figure~\ref{fig:accuracy} presents the classification accuracy 
for the evaluated descriptors. Notice that the classification 
accuracies are low for all the descriptors, creating a very 
challenging scenario for the BoG model, as we explained previously.
The best results were obtained for the SVM model learned on 
800 training frames per class. This model was used for representing 
the test videos using the BoG approach in the following experiments.

\begin{figure*}[!htb]
	\centering
	\subfigure[MAP]{
		\includegraphics[width=0.45\textwidth]{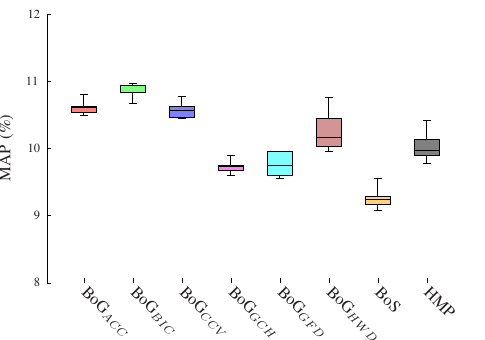}
	}
	\qquad
	\subfigure[P@10]{
		\includegraphics[width=0.45\textwidth]{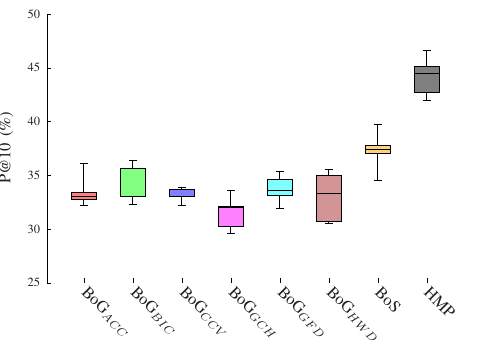}
	}
	\caption{Results for video genre retrieval comparing BoG with 
	         the baselines in terms of MAP and P@10. 
	         BoG$_{BIC}$ obtained the best MAP score.}
	\label{fig:overall}
\end{figure*}

\subsubsection{Evaluation of the BoG representation}
\label{sec:results_2}
The following experiments evaluate the BoG model for video genre retrieval. 
Each video in the test set was represented by a bag of genres 
using the genre classifiers learned on the training step. With 
the BoG of each video, a given test video was used as query for 
the rest of the videos in the test set, which were ranked according 
to the Euclidean ($L_2$) distance between their BoGs.
For each genre, around five percent of the test videos were 
randomly selected and used as queries. Five replications were 
performed in order to ensure statistically sound results. 
Presented results refer to the average scores and their 
respective 99\% confidence intervals, which were computed 
based on the mean and standard deviation of each replication. 

We compared the BoG approach against with two baselines: 
Histogram of Motion Patterns~(HMP)~\cite{ICIP_2011_Almeida} 
and Bag of Scenes~(BoS)~\cite{ICMR_2012_Penatti}. To make a 
fair comparison, these approaches were configured with their 
best settings based on the results reported in~\cite{CIARP_2014_Almeida}.
The distance function used for feature comparison is the 
Euclidean ($L_2$) distance. The retrieval effectiveness 
was assessed using the precision at the top 10 retrieved 
items (P@10) and Mean Average Precision (MAP).

In Figure~\ref{fig:overall}, we compare the BoG representations 
and the baseline methods with respect to the MAP and P@10 measures. 
As we can observe, the performance of the BoG representations are 
slightly better considering the MAP measure. MAP is a good indication 
of the effectiveness considering all positions of obtained ranked lists. 
P@10, in turn, focuses on the effectiveness of the methods considering 
only the first positions of the ranked lists.

The BoG approach achieved the best scores using BIC as the 
frame descriptor (used as basis for the genre classifier). 
Notice that BoG$_{BIC}$ performs better than the baseline 
methods for MAP, however the same does not happen for P@10. 
BIC was the best descriptor for the genre classifier in 
the test set (see Section~\ref{sec:results_1}), making it 
also better for generating the BoG vector. 

\begin{figure*}[!htb]
	\centering
	\subfigure[MAP]{
		\includegraphics[width=0.45\textwidth]{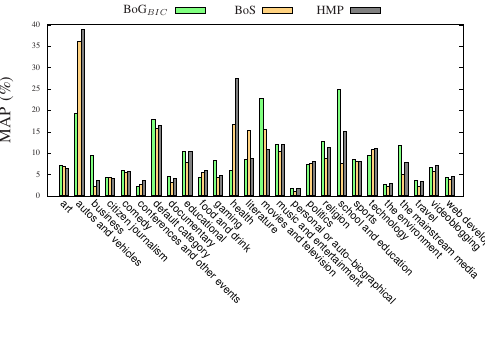}
	}
	\qquad
	\subfigure[P@10]{
		\includegraphics[width=0.45\textwidth]{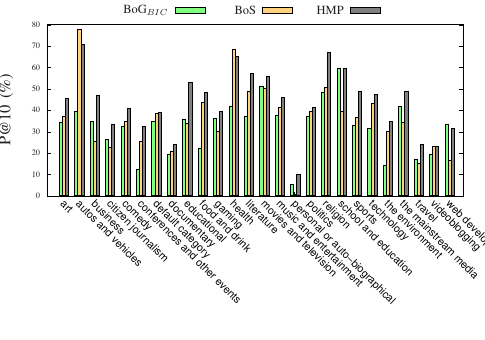}
	}
	\caption{MAP and P@10 scores obtained for each genre.}
	\label{fig:perclass}
\end{figure*}

We also performed paired $t$-tests to verify the statistical 
significance of the results. For that, the confidence intervals for 
the differences between paired averages of each class were computed 
to compare every pair of approaches. If the confidence interval 
includes zero, the difference is not significant at that confidence 
level. If the confidence interval does not include zero, then the 
sign of the difference indicates which alternative is better. 

Table~\ref{tab:confidence} presents the 99\% confidence intervals 
of the differences between BoG$_{BIC}$ (the best configuration of 
BoG) and the baseline methods for the MAP and P@10 measures, respectively. 
Notice that the confidence intervals for BoG$_{BIC}$ and BoS 
include zero and, hence, the differences between those approaches 
are not significant at that confidence level. On the other hand, 
the performance of BoG$_{BIC}$ and HMP are not statistically 
different for MAP, whereas BoG$_{BIC}$ performs worse than HMP 
for P@10. This method is based on motion information and, hence, 
it does not consider visual properties of video frames in an 
independent manner.

\begin{table}[!htb]
	\centering
	\scriptsize
	\caption{Paired $t$-test comparing the best BoG configuration 
	         and the baselines. We can note intervals crossing the 
	         zero for BoG$_{BIC}$ and BoS, indicating no statistical 
	         difference between methods. For BoG$_{BIC}$ versus HMP, 
	         HMP is better for P@10.}
	\begin{tabular}{|c|c|c|c|c|}
		\hline
		\hline
		& \multicolumn{2}{c|}{\textbf{MAP}} 
		& \multicolumn{2}{c|}{\textbf{P@10}} \\
		\cline{2-5}
		\up{\textbf{Approach}} 
		& \makebox[12mm][c]{\textbf{min.}} & \makebox[12mm][c]{\textbf{max.}}
		& \makebox[12mm][c]{\textbf{min.}} & \makebox[12mm][c]{\textbf{max.}} \\
		\hline
		BoG$_{BIC}$ - BoS & \textcolor{red}{-0.018} & 0.018 & \textcolor{red}{-0.063} & 0.014 \\
		BoG$_{BIC}$ - HMP & \textcolor{red}{-0.074} & 0.007 & \textcolor{red}{-0.232} & \textcolor{red}{-0.079} \\
		\hline
		\hline
	\end{tabular}
	\label{tab:confidence}
\end{table}

Figure~\ref{fig:perclass} compares the individual scores 
obtained for each class in terms of MAP and P@10 measures. 
It is interesting to note the differences in responsiveness of 
the different approaches with respect to each of the genres. For 
MAP, BoG$_{BIC}$ performs better than the baseline methods for 
most of the classes (13 out of 26). For P@10, BoG$_{BIC}$ provides 
a good discriminative power on genres like \textit{``school 
and education''} and \textit{``web development and sites''}.

The key advantage of the BoG model is its computational efficiency 
in terms of space occupation and similarity computation time. 
In our experiments, the BoG vector corresponds to a 26-bin histogram, 
which represents a reduction of 74\% in relation to the BoS vector 
(100-bin histogram) and is two orders of magnitude smaller than 
the HMP vector (6075-bin histogram), making our approach more 
suitable for real-time processing.

\begin{table*}[!htb]
\centering
\caption{EVVE events list. The dataset has a total of 620 query videos and 2,375 database videos divided into 13 events. Q refers to the number of queries, Db+ and Db- are the numbers of positive and negative videos in the database, respectively.}
\label{tab:evve-event-list}
\begin{tabular}{c|l|c|c|c}
\hline
\textbf{ID}  & \textbf{Event name}                                & \textbf{Q} & \textbf{Db+}  & \textbf{Db-}  \\ \hline
1            & Austerity riots in Barcelona, 2012                 & 13               & 27            & 122           \\ 
2            & Concert of Die toten Hosen, Rock am Ring, 2012     & 32               & 64            & 143           \\ 
3            & Arrest of Dominique Strauss-Kahn                   & 9                & 19            & 60            \\ 
4            & Egyptian revolution: Tahrir Square demonstrations  & 36               & 72            & 27            \\ 
5            & Concert of Johnny Hallyday stade de France, 2012   & 87               & 174           & 227           \\ 
6            & Wedding of Prince William and Kate Middleton       & 44               & 88            & 100           \\ 
7            & Bomb attack in the main square of Marrakech, 2011  & 4                & 10            & 100           \\ 
8            & Concert of Madonna in Rome, 2012                   & 51               & 104           & 67            \\ 
9            & Presidential victory speech of Barack Obama 2008   & 14               & 29            & 56            \\ 
10           & Concert of Shakira in Kiev 2011                    & 19               & 39            & 135           \\ 
11           & Eruption of Strokkur geyser in Iceland             & 215              & 431           & 67            \\ 
12           & Major autumn flood in Thailand, 2011               & 73               & 148           & 9             \\ 
13           & Jurassic Park ride in Universal Studios theme park & 23               & 47            & 10            \\ \hline
\textbf{All} & \textbf{\textgreater\textgreater\textgreater}      & \textbf{620}     & \textbf{1252} & \textbf{1123} \\ \hline
\end{tabular}
\end{table*}

Although the effectiveness the BoG approach is not superior to the 
baseline methods, the obtained results show the potential of the idea. 
As we explained previously, the quality of the genre classifier 
is important for the BoG quality. Our genre classifiers obtained 
less than 50\% of accuracy in the training set and less than 30\% 
in the test set, probably limiting the quality of the BoG representation.
Another limitation is the dataset used. As all the frames of a video 
have the same label, visually different frames may be of the same 
genre, harming the classifier.

\subsection{Video Event Retrieval}
\label{sec:results_3}
Also, we carried out this study on the EVVE (EVent VidEo) 
dataset\footnote{\url{http://pascal.inrialpes.fr/data/evve/} 
(As of May 2016).}: an event retrieval benchmark introduced 
by Revaud~et~al.~\cite{CVPR_2013_REVAUD}. The dataset is 
composed of 2,995 videos (166 hours) collected from 
YouTube\footnote{\url{http://www.youtube.com} (As of May 
2016).}. Those videos are distributed among 13 event categories 
and are divided into a query set of 620 (20\%) videos and a 
reference collection of 2,375 (80\%) videos. Each event is 
treated as an independent subset containing some specific 
videos to be used as queries and the rest to be used as database 
for retrieval, as shown in Table~\ref{tab:evve-event-list}. 
It is a challenging benchmark since the events are localized 
in both time and space, for instance, the event 1 refers to 
the great riots and strikes that happened in the end of 
March 2012 at Barcelona, Spain, however, in the database, 
there are a lot of videos from different strikes and riots 
around the world.

EVVE uses a standard retrieval protocol: a query video is 
submitted to the system which returns a ranked list of similar 
videos. Then, we evaluate the average precision~(AP) of each 
query and compute the mean average precision~(mAP) per event.
The overall performance is assessed by the average of the 
mAPs~(avg-mAP) obtained for all the events. 

Our experiments followed the official experimental protocol created 
by~\cite{CVPR_2013_REVAUD}. Initially, each video in the dataset was 
represented by a BoG. With the BoG of each video, a given query video 
was used to retrieve the rest database videos, which were ranked 
according to the Euclidean (L2) distance between their BoAs. Finally, 
we used the dataset official tool to evaluate the retrieval 
results\footnote{\url{http://pascal.inrialpes.fr/data/evve/eval_evve.py} 
(As of May 2016).}.

In this experiment, we used BoG$_{BIC}$ to represent videos, 
which was the approach that achieved the best scores for video 
genre retrieval. Our intend here is to verify if the BoG 
representation can perform well in a different scenario. 

We compared the BoG$_{BIC}$ approach against three 
baselines~\cite{CVPR_2013_REVAUD}: Mean-MultiVLAD (MMV), 
CTE (Circulant Temporal Encoding) and a combination of 
both methods, known as MMV+CTE. Also, we considered the 
variations of MMV with the following hyper-pooling 
functions~\cite{ICCV_2013_DOUZE}: k-means, partial 
k-means (PKM), sign of stable componentes (SSC), 
KD-Tree and Fisher Vectors. To make a fair comparison, 
these approaches were selected with their best 
performance based on the results reported 
in~\cite{CVPR_2013_REVAUD, ICCV_2013_DOUZE}. 

\begin{figure}[!htb]
\centering
\includegraphics[width=0.45\textwidth]{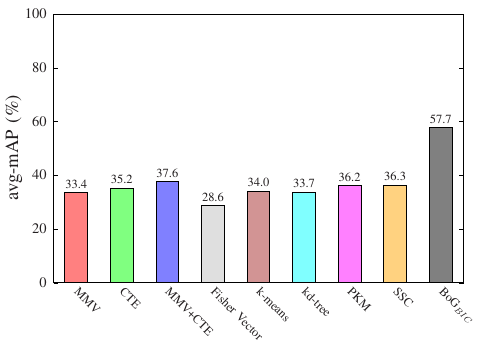}
\caption{Performance of different methods for event retrieval on EVVE dataset.}
\label{fig:avg-map-all}
\end{figure}

In Figure~\ref{fig:avg-map-all}, we compare the BoG$_{BIC}$ 
representation and the baseline methods with respect to the 
avg-mAP. As wecan observe, the performance of the BoG$_{BIC}$ 
representation outperformed all baseline methods by a large margin.

The results where also compared by event, as shown in
Table~\ref{tab:avg-map-bog-event}. One can notice that 
BoG$_{BIC}$ representation performed better than the 
baseline methods for most of the events (11 out of 13).
For some events, the difference in favor of our method 
is very large, like in events 4, 5, 8, and 12. 

\begin{table}[!htb]
\centering
\caption{Retrieval performance (mAP) per event on EVVE dataset.}
\label{tab:avg-map-bog-event}
\begin{tabular}{c|c|c|c|c}
\hline
\textbf{Event ID} & \textbf{MMV} & \textbf{CTE}   & \textbf{MMV+CTE} & \textbf{BoG$_{BIC}$} \\ \hline

1                 & 23.90        & 13.90          & 24.60            & \textbf{25.32}        \\ 
2                 & 19.90        & 16.60          & 20.20            & \textbf{42.58}        \\ 
3                 & 8.70         & 12.80          & 11.10            & \textbf{45.51}        \\ 
4                 & 12.60        & 10.80          & 13.20            & \textbf{79.41}        \\ 
5                 & 23.40        & 26.20          & 26.00            & \textbf{47.20}        \\ 
6                 & 33.80        & 41.30          & 39.40            & \textbf{56.66}        \\ 
7                 & 12.40        & 25.20          & 21.20            & \textbf{33.63}        \\ 
8                 & 25.40        & 25.70          & 28.10            & \textbf{74.71}        \\ 
9                 & 53.10        & \textbf{80.30} & 69.40            &         28.05         \\ 
10                & 45.50        & 40.90          & \textbf{48.60}   &         45.11         \\ 
11                & 77.30        & 71.40          & 77.40            & \textbf{89.04}        \\ 
12                & 36.60        & 29.70          & 37.10            & \textbf{98.57}        \\ 
13                & 60.40        & 69.30          & 71.90            & \textbf{84.87}        \\ \hline
avg-mAP           & 33.40        & 35.20          & 37.60            & \textbf{57.74}        \\ \hline
\end{tabular}
\end{table}

We also performed paired $t$-tests to verify the statistical significance
of the results. For that, the confidence intervals for the differences
between paired averages (mAP) of each category were computed to compare
every pair of approaches.

Table~\ref{tab:paired-t-test} presents the 95\% confidence intervals of 
the differences between BoG$_{BIC}$ and the baseline methods for the mAP measures. 
Notice that the confidence intervals for BoG$_{BIC}$ and the baseline methods 
are always positive, indicating that BoG$_{BIC}$ outperformed those approaches.

\begin{table}[!htb]
	\centering
	\scriptsize
	\caption{Paired t-test comparing BoG$_{BIC}$ and the baselines. As intervals are above zero, we can say that BoG$_{BIC}$ outperformed the baselines with statistical significance.}
\label{tab:paired-t-test}
	\begin{tabular}{|c|c|c|c|}
		\hline
		\hline
		& \multicolumn{2}{c|}{\textbf{mAP}} \\
		\cline{2-3}
		\up{\textbf{Approach}} 
		& \makebox[12mm][c]{\textbf{min.}} & \makebox[12mm][c]{\textbf{max.}} \\
		\hline
		BoG$_{BIC}$ - MMV         & 0.091   & 0.398   \\
		BoG$_{BIC}$ - CTE         & 0.034   & 0.407   \\
		BoG$_{BIC}$ - MMV+CTE     & 0.031   & 0.373   \\
		\hline
		\hline
	\end{tabular}
	\label{tab:paired-t-test}
\end{table}

According to the analysis of BoG$_{BIC}$ results per event, one of 
the worst results happened on the event 1. On the other side, the 
best event was obtained on the event 13. We made a visual analysis 
at the videos to help to understand the differences.

\begin{figure*}[!htb]
	\centering
	\subfigure[Positive Videos]{
		\includegraphics[width=0.45\textwidth]{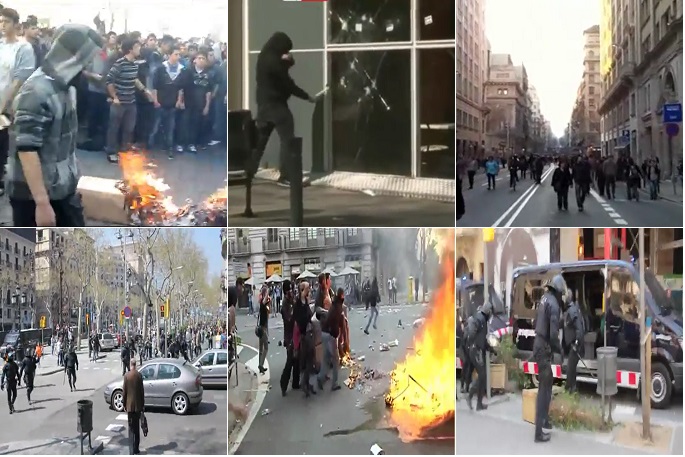}
		\label{fig:barcelona-yes-all}
	}
	\qquad
	\subfigure[Negative Videos]{
		\includegraphics[width=0.45\textwidth]{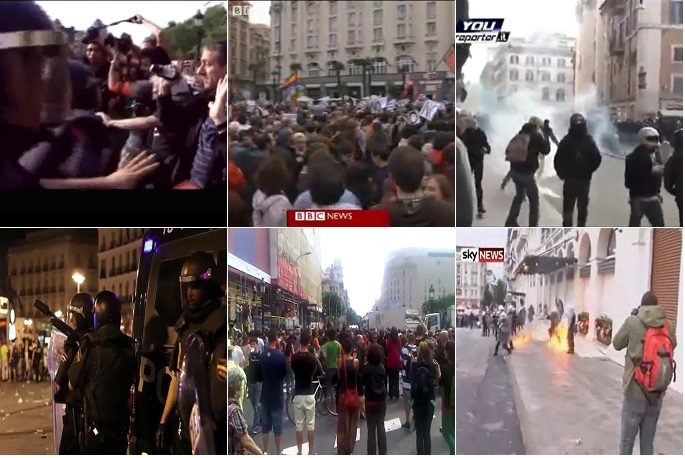}
		\label{fig:barcelona-not-all}
	}
	\caption{Examples of video frames from Event 1 (Austerity riots in Barcelona, 2012), which was one of the events that BoG performed worst.}
	\label{fig:barcelona}
\end{figure*}

In case of the event 1 (see Figure~\ref{fig:barcelona}), it 
is possible to see lots of riots and strikes at different places 
and moments. There are scenes showing police, fire, cars, and crowd 
in almost all the videos (Figure~\ref{fig:barcelona-not-all}). 
Thus, it is difficult to identify only videos of the austerity riots that 
occurred in Barcelona at the end of March, 2012 (Figure~\ref{fig:barcelona-yes-all}). 
As shown in Table~\ref{tab:avg-map-bog-event}, all the methods 
performed below 25\% for this event.

\begin{figure*}[!htb]
	\centering
	\subfigure[Positive Videos]{
		\includegraphics[width=0.45\textwidth]{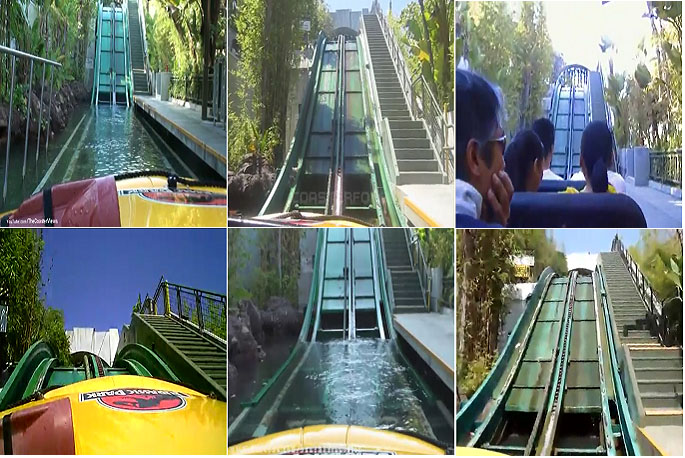}
		\label{fig:jurassic-park-yes-all}
	}
	\qquad
	\subfigure[Negative Videos]{
		\includegraphics[width=0.45\textwidth]{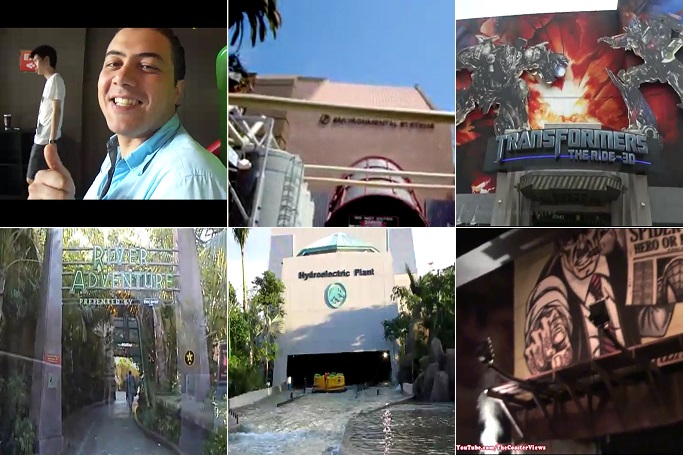}
		\label{fig:jurassic-park-not-all}
	}
	\caption{Examples of video frames from Event 13 (Jurassic Park ride in Universal Studios theme park), which was one of the events that BoG performed best.}
	\label{fig:jurassic}
\end{figure*}

But, in case of the event 13 (see Figure~\ref{fig:jurassic}), there 
are lots of similar positive videos, specially recorded at the entrance 
of the ride, as shown in Figure~\ref{fig:jurassic-park-yes-all}. 
This scene is repeated in many videos and probably helped our method. 
Negative videos do not contain the same entrance, as shown in 
Figure~\ref{fig:jurassic-park-not-all}. 

We believe that our method outperformed the baseline methods because 
the proposed BoG representation carries semantic information. But, on 
the other side, our method does not include temporal information and 
we think such feature is important to recognize some types of events. 

\section{Conclusions}
\label{sec:conclusions}
In this paper, we presented a new video representation for video  
retrieval, named as Bag of Genres. This representation model 
relies on a dictionary of genres, which is created from a genre 
classification model learned on the training frames. Different from 
traditional dictionaries based on local features (e.g., SIFT or 
STIP), here, visual words correspond genre-labeled regions of the 
classification space. Therefore, each dimension of the feature 
space spanned by such a model is associated to a semantic concept. 

Our approach was validated in the dataset of MediaEval Tagging 
Task of 2012. Our experiments evaluated the importance of the genre 
classifier in the model as well as the quality of the BoG representation. 
In these experiments, the BoG model has performed well despite the 
low accuracy of the genre classifier. The results demonstrated that 
our approach performs similar to state-of-the-art methods, but using 
a much more compact representation. 
Also, we tested the best configuration of the BoG model to retrieve 
videos by event on the EVVE dataset. The results show that our approach 
outperformed state-of-the-art solutions. 

We can think about ways of improving the BoG model. For instance, 
a smarter strategy for feature extraction and classification may 
enable to create more informative visual dictionaries and, hence, 
improve the video representation.

Future work includes the evaluation of other methods for feature 
extraction, as well as perform an extensive study on classification 
strategies to be used in the creation of visual dictionaries. We 
also would like to evaluate the use of a dataset of scene images 
to create the genre classifier.

\iffinal
\section*{Acknowledgment}
The authors would like to thank CAPES, CNPq, and 
FAPESP~(grant~\#2016/06441-7) for funding.
\fi



\balance


\end{document}